\documentclass[10pt,twocolumn,letterpaper]{article}

\usepackage[pagenumbers]{iccv} 

\usepackage{xcolor}
\definecolor{lightblue}{RGB}{173,216,230}
\definecolor{lightgreen}{RGB}{144,238,144}
\usepackage{algorithm}
\usepackage{algpseudocode}
\usepackage{multirow}
\usepackage{multicol}
\usepackage{mathtools}
\usepackage{comment}
\usepackage{soul}
\usepackage{arydshln}
\usepackage{pgffor}   
\usepackage{float}    

\definecolor{iccvblue}{rgb}{0.21,0.49,0.74}
\usepackage[pagebackref,breaklinks,colorlinks,allcolors=iccvblue]{hyperref}

\def\paperID{8470} 
\def\confName{ICCV}
\def\confYear{2025}

\title{Zero-Shot Low Light Image Enhancement with Diffusion Prior}

\author{Joshua Cho
\quad\quad
Sara Aghajanzadeh
\quad\quad
Zhen Zhu
\quad\quad
D. A. Forsyth \\
University of Illinois Urbana-Champaign \\
{\tt\small \{joshua66, saraa5, zhenzhu4, daf\}@illinois.edu}
}

\begin{document}

\twocolumn[{
\maketitle
\begin{center}
    \centering
    \captionsetup{type=figure}
    \vspace{-0.3cm}
    \includegraphics[width=\textwidth]{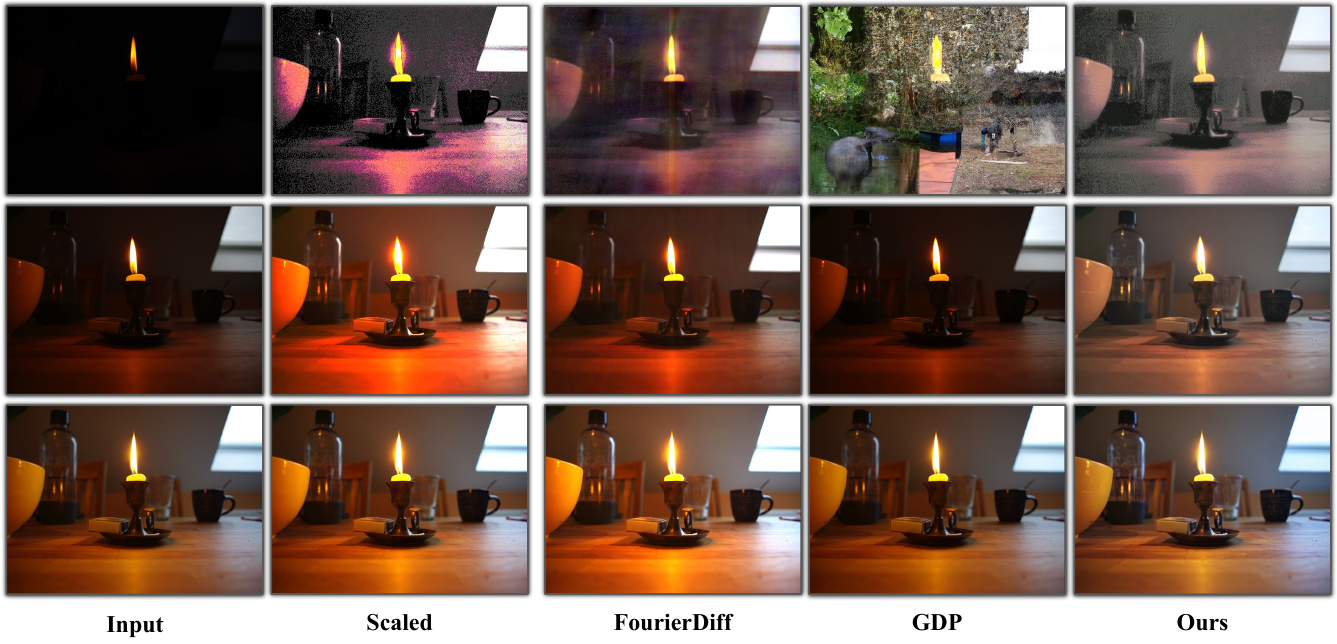}
    \captionof{figure}{In low-light image enhancement, an ideal method should achieve \textbf{color constancy}~\cite{color_constancy_algorithm, color_constancy_algorithm2} by accurately recovering the intrinsic color (reflectance) of a scene, ensuring consistency across images taken under varying illumination conditions. However, suboptimal illumination introduces noise and hampers the accurate capture of all wavelengths, leading to color distortions. For instance, recent zero-shot diffusion-based methods such as GDP~\cite{gendiffprior} suffer from \textbf{hallucinations}, introducing non-existent elements (row 1), while FourierDiff~\cite{FourierDiff} is compromised by the inherent noise sensitivity of frequency-domain representations (rows 1 and 2). In contrast, our approach demonstrates superior color constancy and fidelity across images of the same scene, effectively mitigating the challenges posed by lighting variations.
    }
    \label{fig:teaser}
\end{center}
}]

\begin{abstract}
In this paper, we present a simple yet highly effective ``free lunch'' solution for low-light image enhancement (LLIE), which aims to restore low-light images as if acquired in well-illuminated environments. Our method necessitates no optimization, training, fine-tuning, text conditioning, or hyperparameter adjustments, yet it consistently reconstructs low-light images with superior fidelity. Specifically, we leverage a pre-trained text-to-image diffusion prior, learned from training on a large collection of natural images, and the features present in the model itself to guide the inference, in contrast to existing methods that depend on customized constraints. Comprehensive quantitative evaluations demonstrate that our approach outperforms SOTA methods on established datasets, while qualitative analyses indicate enhanced color accuracy and the rectification of subtle chromatic deviations. Furthermore, additional experiments reveal that our method, without any modifications, achieves SOTA-comparable performance in the auto white balance (AWB) task.
\end{abstract}    
\section{Introduction}

\begin{figure}[t]
  \centering
  \includegraphics[width=\columnwidth]{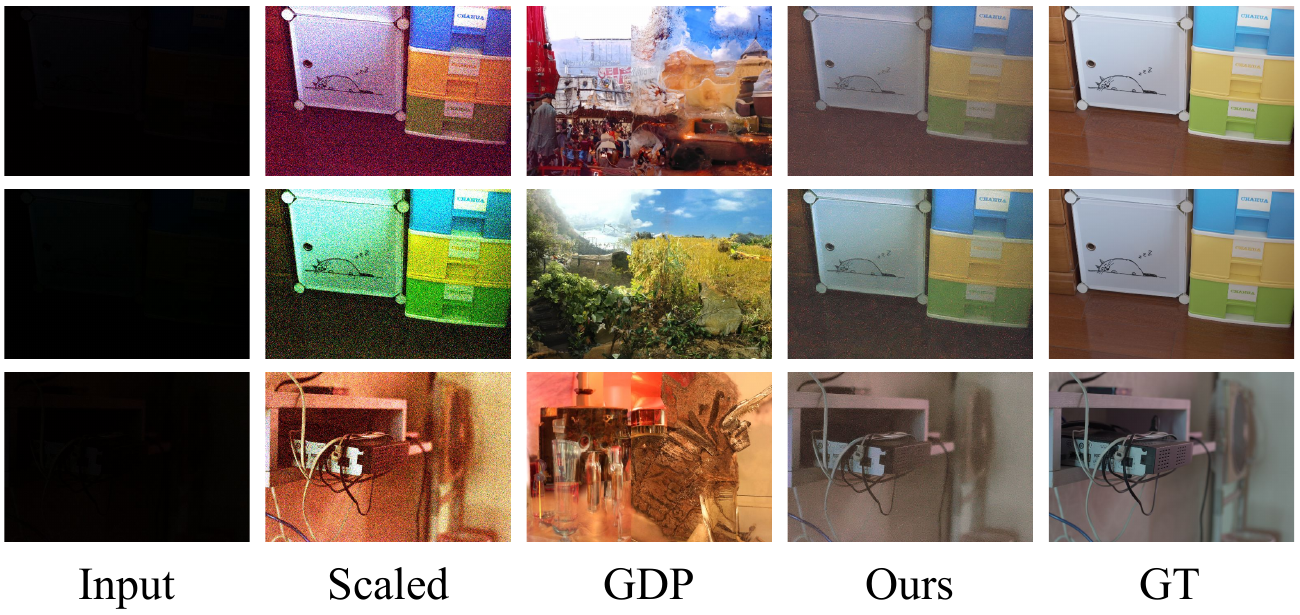}
  \caption{\textbf{Hallucination.} While diffusion prior is effective for image restoration, improper application can lead to unintended hallucinations, where the model generates nonexistent structures or alters scene semantics. For example, GDP~\cite{gendiffprior}, a robust and versatile image restoration method, often hallucinates in the presence of substantial noise and darkness in input images. As shown in row 1, a blue-colored cabinet is inaccurately reconstructed as a sky, a pink cabinet as a building, and the entire scene resembles a battle. For less noisy inputs, our method produces clean and sharp outputs and effectively attenuates noise even in challenging cases involving severe darkness and pronounced noise levels.}
  \label{fig:hallucination}
\end{figure}

\begin{figure*}[ht]
\centering
\includegraphics[width=\textwidth]{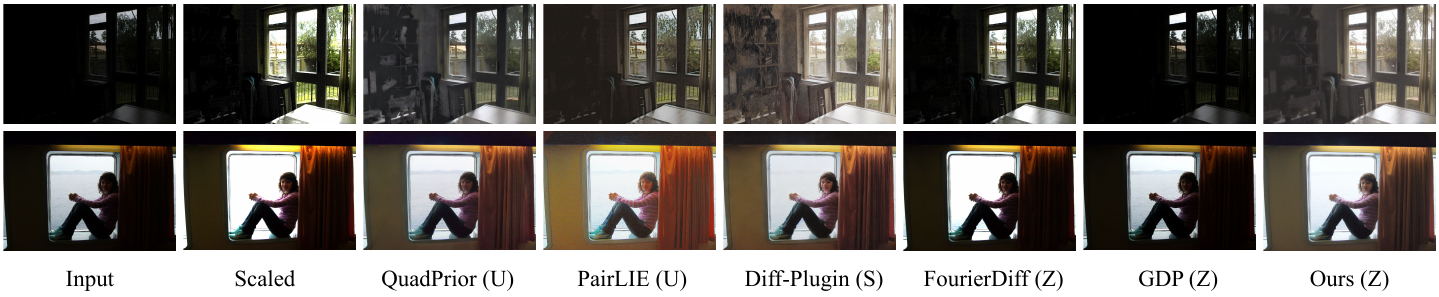} 
\caption{Qualitative evaluation of our method against existing unsupervised, supervised, and zero-shot approaches on the Unpaired dataset. Please \textbf{zoom in without night-light mode} to accurately compare colors and observe noise reduction in each method. For example, PairLIE exhibits color shifts (row 2), while QuadPrior and Diff-Plugin introduce structural distortions in the bookshelf and wooden chair (row 1).}
\label{fig:qual-unpaired}
\end{figure*}

Low-light image enhancement aims to enhance images captured in suboptimal lighting conditions into their natural, well-lit counterpart. Its relevance spans from photography~\cite{mobile-photo} to various downstream tasks such as autonomous driving \cite{autodriving1}, underwater image enhancement \cite{underwater1, underwater2}, and video surveillance \cite{videosurveilance1}. Yet, it is a challenging task because of the presence of shot noise and color quantization effects, which undermine the applicability of elementary solutions such as uniform intensity scaling as evidenced in Figure~\ref{fig:teaser} (first row, Scaled column). Despite advancements in prior studies (discussed in Section~\ref{sec:relatedwork}), the inherent dataset dependencies in both supervised and unsupervised learning paradigms continue to introduce unintended artifacts, as evidenced in Figures~\ref{fig:qual-paired} and ~\ref{fig:qual-unpaired}.

\begin{figure}[t]
    \centering
    \includegraphics[width=\columnwidth]{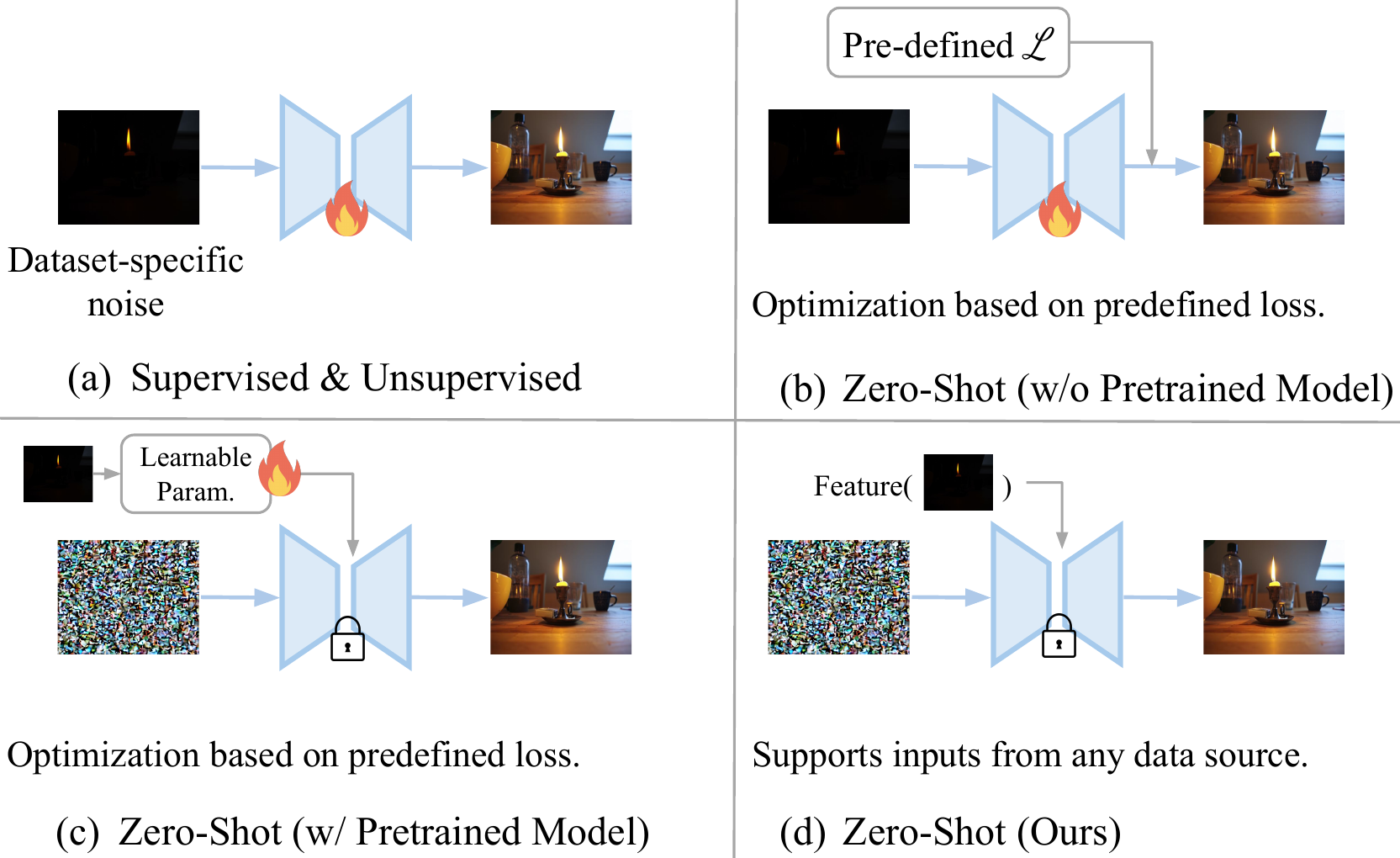}
    \caption{\textbf{LLIE Method Taxonomy.}  \textbf{(category a)} Image regression methods often produce results that are heavily dependent on the dataset, because real-world datasets are small in size.  
      \textbf{(category b)} Zero-shot methods w/o a pre-trained model dynamically adjust model weights per image based on a predefined loss function. However, they require per-image tuning and may suffer from convergence instability.
      \textbf{(category c)} Zero-shot method w/ a pre-trained model and an auxiliary trainable network or parameters that learn on a per-image basis. While this enhances adaptability, it still requires per-image tuning and remains susceptible to convergence instability.  
      \textbf{(category d)} In contrast, our method leverages the self-attention features of pre-trained diffusion models from the input to guide inference
      \textbf{from any data source without any assumption about degradation and without test time tuning.}}
    \label{fig:taxonomy}
\end{figure}

To address this limitation, prior zero-shot LLIE methods~\cite{tao, gendiffprior, FourierDiff} optimize the auxiliary network or parameters alongside a frozen pre-trained diffusion model at test time by relying on custom loss formulation or degradation assumption (Figure~\ref{fig:taxonomy}). Our method, however, leverages internal signals within the model, specifically self-attention features, extending their applicability beyond previous uses in \textit{image editing} tasks~\cite{plug_and_play, masactrl, dragdiffusion, pix2pixzero, StyleAligned, Chung, selfguidance} by following four simple steps: (1) preprocessing; (2) inverting the input image; (3) adjusting the resulting noised latent with Adaptive Instance Normalization (AdaIN) to match standard normal distributions $\mathcal{N}(0, I)$; and (4) denoising the inverted representation with self-attention features extracted during the inversion process.

Qualitative analysis suggests that the robustness of our method arises from the method’s ability to precisely correct subtle color shifts, a significant benefit in low-light images where color degradation is prevalent as shown in Figure~\ref{fig:qual-paired}. Notably, as a secondary outcome, our method also proves effective for white balancing, achieving performance on par with the SOTA approaches.

\textbf{Our key contributions are as follows}:
\begin{enumerate}
    \item We propose a simple yet highly effective zero-shot method for low-light image enhancement (LLIE) that requires no training, fine-tuning, or optimization, yet surpasses state-of-the-art (SOTA) performance on established benchmarks using standard evaluation metrics.
    \item To the best of our knowledge, without any modification, our method is also the first zero-shot auto white balance method and achieves results comparable to SOTA in this task.
\end{enumerate}
\section{Related Work}\label{sec:relatedwork}
\textbf{Traditional Methods.}
Conventional image enhancement methods, including histogram equalization~\cite{histogram1, histogram2} and gamma correction~\cite{gamma}, rely on global adjustments to enhance image contrast.  Despite their computational efficiency, these methods are inherently limited by their inability to account for varying scene-specific lighting conditions. Moreover, their global adjustment inadvertently amplifies dark noise in low-light areas, diminishing fine details and introducing artifacts.

\textbf{Supervised Methods.}
Convolutional Neural Networks (CNNs) are adept at learning transformations from under-exposed to well-lit images, effectively capturing local textures and patterns \cite{cnn1, lolv1, lolv2, excnet}. However, their limitation in capturing long-range dependencies has led to alternative approaches, such as ensemble methods~\cite{aghajanzadeh-equi}, transformer-based method~\cite{retinexformer}, and synthetic data augmentation~\cite{aghajanzadeh, Monakhova}.

Recently, generative methods have exhibited promising results in low-light image enhancement, with diffusion models \cite{diffusionbeats, gradient, ldm, dickstein, ddpm, score} demonstrating particular efficacy because of their strong generative ability, being free from the instability and mode-collapse that are prevalent in previous generative models. However, standard Gaussian noise assumptions of diffusion models do not model the complex noise of low-light images. In response, new training strategies have been proposed for raw and RGB low-light image enhancement~\cite{nguyen, exposurediff, wcdm, lowode, diffretinex, lightendiff, diffplugin}. However, these methods remain dependent on supervised learning, without leveraging the generative priors of the pre-trained diffusion models.

\textbf{Unsupervised Methods.}
Unsupervised learning methods~\cite{RUAS, clip-lit, Zero-DCE, zero-dce++, gan1, sci, clip-lit, nerco, pairlie, zero-ig, quadprior, semantic-guided-llie, lit-the-darkness} have emerged as a promising direction for low-light image enhancement, because they do not rely on paired datasets. EnlightenGAN~\cite{gan1} and NeRCo~\cite{nerco} performs adversarial learning on unpaired data, CLIP-LIT~\cite{clip-lit} leverages CLIP prior and learnable prompt embeddings, and PairLIE~\cite{pairlie} learns priors from paired low-light images. Another category of unsupervised methods is the zero-reference approach, wherein a dataset comprising a single class is leveraged for training. This method capitalizes on the intrinsic color properties of natural images, drawing upon established theoretical frameworks such as Retinex theory~\cite{color_constancy_algorithm2} and the Kubelka-Munk theory~\cite{kubelka-munk}. However, the aforementioned principles may not consistently align with the real-world behavior of noise in under-exposed data. Zero-DCE~\cite{Zero-DCE} and Zero-DCE++~\cite{zero-dce++} employ neural networks to estimate the parameters of a predefined curve function, facilitating adaptive image enhancement. Methods such as RUAS~\cite{RUAS}, SCI~\cite{sci}, and ZeroIG~\cite{zero-ig} employ Retinex-theory-based decomposition to enhance illumination and contrast, whereas QuadPrior~\cite{quadprior}, trained on the COCO~\cite{coco} dataset, relies on the Kubelka-Munk theory. Additionally, Lit-the-Darkness~\cite{lit-the-darkness} and Semantic-GuidedLLIE~\cite{semantic-guided-llie} incorporate custom loss formulation specifically designed to refine color fidelity, texture details, and semantic integrity.

\textbf{Zero-Shot Methods (Previous).}
Existing zero-shot approaches can be broadly classified into three primary categories as shown in Figure~\ref{fig:taxonomy}: \textbf{(category b)} methods that do not rely on a pre-trained model and instead optimize a predefined loss function~\cite{excnet, rrdnet, colie}; \textbf{(category c)} methods with a pre-trained model and an auxiliary trainable network that seeks to minimize a prescribed objective~\cite{gendiffprior, tao, FourierDiff}; and \textbf{(category d)} projection-based methods, which aim to extract intrinsic structures or textures from degraded inputs, thereby guiding inference toward maintaining data fidelity. In (category b), ExCNet~\cite{excnet}, a CNN-based approach, estimates the parametric S-curve at test time by employing a block-based loss function to enhance visibility. RRDNet~\cite{rrdnet} and COLIE~\cite{colie} iteratively minimize a Retinex-based objective to improve image quality. In (category c), GDP~\cite{gendiffprior} approximates the intractable posterior $p(y|x_t)$, for a degraded observation $y$ and its pristine counter-part $x$, through an additional trainable degradation model that is optimized at inference. Meanwhile, TAO~\cite{tao} employs a learnable test-time degradation adapter aimed at minimizing adversarial loss, and FourierDiff~\cite{FourierDiff}, a concurrent work to ours, employs a frequency-domain biasing akin to ILVR~\cite{ilvr} with a learnable brightness parameter and optimizes the phase of the input with a prescribed loss. However, as demonstrated in Table~\ref{tab:quan-metrics}, GDP~\cite{gendiffprior}, TAO~\cite{tao}, and FourierDiff~\cite{FourierDiff} necessitate optimizing learnable parameters per image at inference, incurring significant computational overhead and remains susceptible to convergence instability. Unlike approaches in (category b) and (category c), (category d) offers a compelling advantage because it does not require additional adaptation or optimization, capitalizing on the prior knowledge embedded within a pre-trained model without relying on external prior assumptions or constraints. By modulating inference based on the deep feature representations of the input data, these methods remain independent from the specific degradation assumptions, enabling broad applicability across diverse data sources. In general image restoration tasks, notable works include RePaint~\cite{repaint}, ILVR~\cite{ilvr}, and CCDF~\cite{ccdf}.

\textbf{Auto White Balance (AWB).}
Auto White Balance (AWB) aims to correct color temperature in images for natural color reproduction across diverse lighting conditions. A text-to-image white balancing approach in SDXL~\cite{sdxl} was introduced by~\cite{timothy} and proposed an approach that shifts the mean of each channel toward a specified target value at each denoising step. They suggested that each channel of the latent encoded by the VAE sequentially represents luminance, cyan/red, lime/medium purple, and pattern/structure. While~\cite{timothy} introduced AWB in the text-to-image domain by modifying the mean at every denoising step, to the best of our knowledge, we are the first to propose a zero-shot AWB method that is directly applicable to color-imbalanced images. Without any modifications to the LLIE framework, our approach achieves competitive performance with supervised AWB methods~\cite{mixedillwb, wb_srgb, quasi-cc} while surpassing state-of-the-art image restoration methods~\cite{ddnm, tao}.

\begin{figure*}[t]
  \centering
  \includegraphics[width=\textwidth]{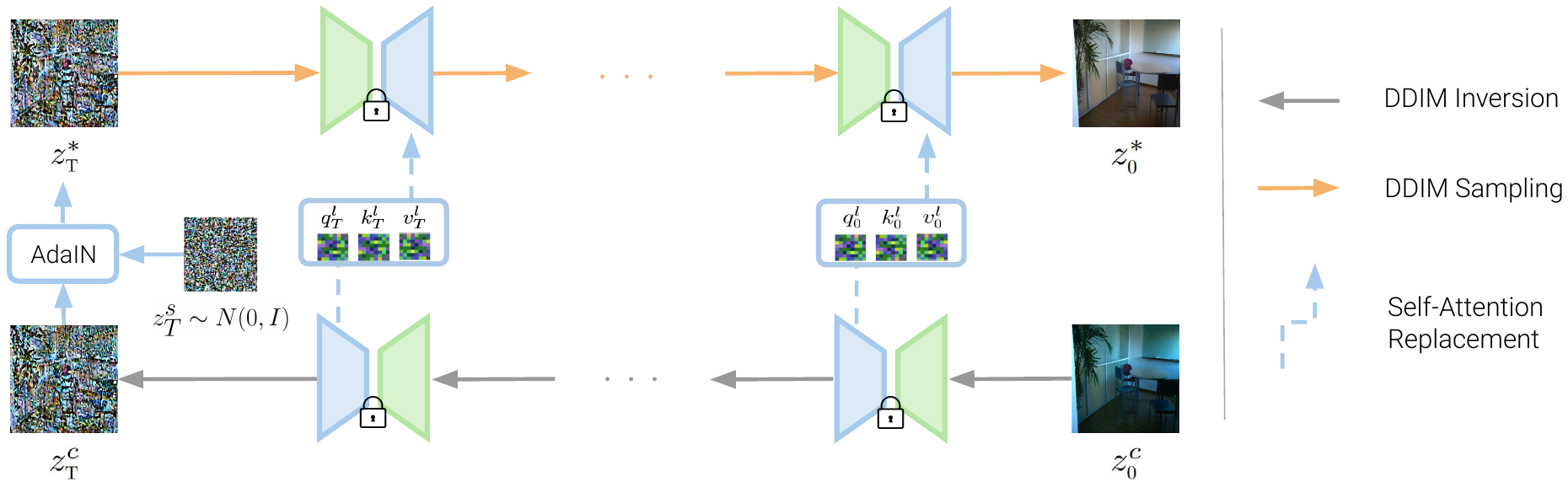}
  \caption{\textbf{Overall pipeline of our method.} Our method offers a simple yet highly effective ``free lunch'' solution for both LLIE and AWB and consists of four main steps: (1) preprocessing; (2) inverting the input image; (3) adjusting the resulting noised latent with Adaptive Instance Normalization (AdaIN) to match standard normal distributions $\mathcal{N}(0, I)$; and (4) denoising the inverted representation with self-attention features extracted during the inversion process, without relying on prior assumptions or external constraints.}
  \label{fig:our-method}
\end{figure*}
\section{Method}\label{sec:method}

Our method for low-light enhancement and auto-white balance consists of four main steps: (1) preprocessing; (2) inverting the input image; (3) adjusting the resulting noised latent with Adaptive Instance Normalization (AdaIN) to match standard normal distributions $\mathcal{N}(0, I)$; and (4) denoising the inverted representation with self-attention features extracted during the inversion process.

\textbf{Preprocessing.}
If the original intensity of the input image \( \textit{I} \in [0, 255] \) falls below a threshold of 30.0, we upscale its average intensity to this level; otherwise, we do not change the input. No other additional pre-processing or post-processing is introduced.

\textbf{Inversion.}
The VAE produces a latent code \( z^c_0 \) from the preprocessed input. We apply DDIM-inversion~\cite{ddim} to obtain its corresponding noisy state \( z^c_T \) after \( T \) timesteps ($T = 25$). During DDIM-inversion, we also extract and store the intermediate self-attention features $\{q^{l}_t, k^{l}_t, v^{l}_t\}$ from each layer \( l \) of the up block layers in the Stable Diffusion UNet across all timesteps \( t \in \{1, \ldots, T\} \).

\textbf{Normalization.}
We apply Adaptive Instance Normalization (AdaIN)~\cite{adain} to the inverted state:
\begin{equation}
    \mathbf{z}^*_T 
    \;=\;
    \sigma\bigl(\mathbf{z}^s_T\bigr)\;
    \left(\frac{\mathbf{z}^c_T \; - \; \mu\bigl(\mathbf{z}^c_T\bigr)}{\sigma\bigl(\mathbf{z}^c_T\bigr)}\right)
    \;+\;
    \mu\bigl(\mathbf{z}^s_T\bigr),
\end{equation}%
\noindent where $\mu(\cdot)$ and $\sigma(\cdot)$ denote the channel-wise mean and standard deviation, respectively, and $\mathbf{z}^s_T \sim \mathcal{N}(0, \mathbf{I})$. 

\textbf{Denoising.}
As direct DDIM denoising would cause drift as shown in Figure~\ref{fig:ablation-images} (AdaIN + DDIM; and DDIM), we replace the default self-attention features formed when denoising \( z^*_T \), with the previously extracted self-attention features from the input, $\{q^{l}_t, k^{l}_t, v^{l}_t\}$, in the corresponding up blocks of the model.

\sethlcolor{lightblue}
\setlength\dashlinedash{0.2em} 
\setlength\dashlinegap{0.2em}  

\begin{table*}[ht]
\centering
\scriptsize
\setlength{\tabcolsep}{2pt}
\renewcommand{\arraystretch}{1.1}
\begin{tabular}{c|l|c|c|c|c|ccc|ccc|ccc}
\toprule
& \multirow{2}{*}{Method} & \multirow{2}{*}{\centering Train Data} & \multirow{2}{*}{\centering Time} & \multirow{2}{*}{\centering Memory} & \multirow{2}{*}{\centering FLOPs} & \multicolumn{3}{c|}{LOL} & \multicolumn{3}{c|}{LSRW} & \multicolumn{3}{c}{Unpaired} \\
\cmidrule(lr){7-9} \cmidrule(lr){10-12} \cmidrule(lr){13-15}
& & & & & & PSNR $\uparrow$ & SSIM $\uparrow$ & LPIPS $\downarrow$ & PSNR $\uparrow$ & SSIM $\uparrow$ & LPIPS $\downarrow$ & ILNIQE $\downarrow$ & BRISQUE $\downarrow$ & NL $\downarrow$ \\
\midrule

\multirow{10}{*}{S} 
& KinD~\cite{kind} \textit{MM'19} & LOLv1 & 0.06 & 1.8 & 1.6e+07 & 20.2047 & 0.8140 & 0.1475 & 16.4070 & 0.4841 & 0.3371 & 27.3155 & 36.1660 & 0.4650 \\
& KinD++~\cite{kind++} \textit{IJCV'21} & LOLv1 & 0.24 & 1.4 &  1.7e+07 & 17.6722 & 0.7691 & 0.2140 & 16.0854 & 0.4025 & 0.3704 & 25.3412 & 34.1082 & 0.5698 \\
& SNR~\cite{snr} \textit{CVPR'22} & LOLv1 & 0.02 & 4.2 & 1.9e+11 & \textbf{21.8877} & \textbf{0.8480} & 0.1561 & 17.8400 & \textbf{0.5639} & 0.4774 & 28.4187 & 35.2478 & 0.4037 \\
& GSAD~\cite{GSAD} \textit{NeurIPS'23} & LOLv1 & 0.07 & 9.3 & 1.1e+13 & 20.6046 & 0.8470 & \textbf{0.1118} & 17.5835 & 0.5509 & 0.3269 & 26.3613 & 29.0197 & 0.7501 \\
& Retinexformer~\cite{retinexformer} \textit{ICCV'23} & MIT5K & 0.02 & 7.1 & 1.2e+11 & 13.0263 & 0.4256 & 0.3649 & 11.4123 & 0.2689 & 0.5060 & 36.3183 & 38.2946 & \textbf{0.3994} \\
& Diff-Plugin~\cite{diffplugin} \textit{CVPR'24} & LOLv1 & 0.09 & 8.0 & 2.9e+13 & 18.8273 & 0.7041 & 0.1826 & \textbf{17.9620} & 0.5224 & \textbf{0.3032} & \textbf{22.6731} & \textbf{18.1443} & 0.8530 \\
\midrule

\multirow{17}{*}{U} 
& Zero-DCE~\cite{Zero-DCE} \textit{CVPR'20} & own data & 0.03 & 1.8 & 3.8e+10 & 17.6417 & 0.5717 & 0.3154 & 15.8680 & 0.4506 & \textbf{0.3152} & 26.8335 & 23.1635 & 1.5336 \\
& Zero-DCE++~\cite{zero-dce++} \textit{TPAMI'21} & own data & 0.03 & 4.0 & 4.5e+7 & 17.0175 & 0.4439 & 0.3145 & 16.2453 & 0.4568 & 0.3273 & 25.9503 & \textbf{17.1177} & 1.2573 \\
& EnlightenGAN~\cite{gan1} \textit{TIP'21} & LOLv1+ & 0.02 & 4.4 & 2.4e+11 & 18.4888 & 0.6722 & 0.3105 & 17.0811 & 0.4705 & 0.3273 & 24.7138 & 18.9759 & 0.9667 \\
& RUAS~\cite{RUAS} \textit{CVPR'21} & LOLv1 & 0.02 & 4.1 & 1.6e+9 & 15.4663 & 0.4892 & 0.3045 & 14.2711 & 0.4698 & 0.4650 & 63.0158 & 25.7555 & - \\
& RUAS~\cite{RUAS} \textit{CVPR'21} & DarkFace & 0.02 & 4.1 & 1.6e+9 & 15.0502 & 0.4562 & 0.3716 & 14.0305 & 0.4028 & 0.3847 & 39.2783 & 26.9609 & 3.8925 \\
& RUAS~\cite{RUAS} \textit{CVPR'21} & MIT5K & 0.02 & 4.1 & 1.6e+9 & 13.6270 & 0.4616 & 0.3464 & 13.0235 & 0.3585 & 0.3790 & 31.2611 & 29.8218 & 2.2491 \\
& SCI~\cite{sci} \textit{CVPR'22} & LOLv1+ & 0.02 & 4.0 & 1.2e+8 & 16.9749 & 0.5320 & 0.3120 & 15.2419 & 0.4240 & 0.3218 & 28.7476 & 24.5364 & 2.0011 \\
& SCI~\cite{sci} \textit{CVPR'22} & DarkFace & 0.02 & 4.0 & 1.2e+8 & 16.8033 & 0.5436 & 0.3225 & 15.1626 & 0.4080 & 0.3259 & 28.0069 & 21.3410 & 1.3388 \\
& SCI~\cite{sci} \textit{CVPR'22} & MIT5K & 0.02 & 4.0 & 1.2e+8 & 11.6632 & 0.3948 & 0.3616 & 11.7939 & 0.3173 & 0.4004 & 27.6531 & 18.0679 & 0.8525 \\
& SemanticGuidedLLIE~\cite{semantic-guided-llie} \textit{WACV'22} & own data & 0.03 & 4.1 & 4.6e+8 & 17.1981 & 0.4419 & 0.3161 & 16.6963 & 0.4577 & 0.3242 & 26.2716 & 17.4838 & 1.3544 \\
& Lit-the-Darkness~\cite{lit-the-darkness} \textit{ICASSP'23} & own data & 0.03 & 4.0 & 4.6e+8 & 18.2337 & 0.5711 & 0.3165 & 16.8203 & 0.4560 & 0.3187 & 27.1650 & 20.1043 & 1.5600 \\
& CLIP-LIT~\cite{clip-lit} \textit{ICCV'23} & own data & 0.02 & 4.0 & 1.3e+11 & 14.8179 & 0.5243 & 0.3706 & 13.4835 & 0.4051 & 0.3533 & 28.1403 & 26.0421 & 1.8638 \\
& NeRCo~\cite{nerco} \textit{ICCV'23} & LOLv1 & 0.03 & 5.0 & 1.7e+12 & \textbf{21.0728} & 0.7248 & 0.2591 & 17.3814 & 0.5303 & 0.5182 & 28.1309 & 30.0167 & 0.8358 \\
& PairLIE~\cite{pairlie} \textit{CVPR'23} & LOLv1+ & 0.02 & 4.1 & 1.6e+11 & 19.6999 & 0.7737 & 0.2353 & \textbf{17.6104} & 0.5190 & 0.3309 & 26.7961 & 31.0421 & 1.64065 \\
& ZeroIG~\cite{zero-ig} \textit{CVPR'24} & LSRW & 0.02 & 4.1 & 7.7e+10 & 17.5677 & 0.4778 & 0.3799 & 16.7516 & 0.5010 & 0.4000 & 34.2056 & 34.8248 & 2.4482 \\
& ZeroIG~\cite{zero-ig} \textit{CVPR'24} & LOLv1 & 0.02 & 4.1 & 7.7e+10 & 18.6589 & 0.7496 & 0.2415 & 16.4431 & 0.5087 & 0.3764 & 27.5778 & 26.2227 & 1.4861 \\
& QuadPrior~\cite{quadprior} \textit{CVPR'24} & COCO & 0.12 & 12.2 & 4.2e+13 & 20.3016 & \textbf{0.8096} & \textbf{0.2032} & 16.9469 & \textbf{0.5601} & 0.3824 & \textbf{24.4373} & 18.1736 & \textbf{0.4215} \\
\hdashline

\multirow{7}{*}{Z} 
& ExCNet~\cite{excnet} \textit{MM'19} \hl{(b)} & n/a & 0.37 & 1.5 & 1.7e+7 & 16.2972 & 0.4589 & 0.3745 & 15.7021 & 0.4098 & 0.3375 & 27.3933 & 19.2416 & 1.8234 \\
& RRDNet~\cite{rrdnet} \textit{ICME'20}  \hl{(b)} & n/a & 0.52 & 22.5 & 6.1e+13 & 13.5719 & 0.4791 & 0.3238 & 13.4272 & 0.3918 & 0.3358 & 26.6567 & 17.9583 & 1.1570 \\
& GDP~\cite{gendiffprior} \textit{CVPR'23}  \hl{(c)} & n/a & 19.09 & 4.7 & - & 14.6630 & 0.5037 & 0.3559 & 13.0678 & 0.3918 & 0.4476 & 28.6436 & 27.0142 & 0.5280 \\
& TAO~\cite{tao} \textit{ICML'24}  \hl{(c)} & n/a & 3.5 & 4.7 & 4.7e+15 & 19.1807 & 0.6065 & 0.3897 & 15.6891 & 0.4306 & 0.7023 & 42.0846 & 42.1383 & 0.3840 \\
& COLIE~\cite{colie} \textit{ECCV'24}  \hl{(b)} & n/a & 0.05 & 4.7 & 5.2e+12 & 14.8999 & 0.4985 & 0.3268 & 14.0013 & 0.4053 & 0.3424 & 26.8185 & 18.9680 & 0.9642 \\
& FourierDiff~\cite{FourierDiff} \textit{CVPR'24}  \hl{(c)} & n/a & 0.82 & 7.1 & 8.5e+14 & 16.9525 & 0.6039 & 0.2934 & 15.6251 & 0.4610 & 0.3207 & 25.9272 & 26.5694 & 1.2206 \\
& Ours  \hl{(d)} & n/a & 0.12 & 6.7 & 5.8e+13 & \textbf{21.7393} & \textbf{0.8152} & \textbf{0.1771} & \textbf{17.6634} & \textbf{0.5185} & \textbf{0.2829} & \textbf{25.4181} & \textbf{16.1843} & \textbf{0.3794} \\
\bottomrule
\end{tabular}
\caption{Qualitative comparison on the widely used datasets: LOL, LSRW, and Unpaired. We denote LOLv1+ as a dataset collection comprising LOL and other datasets. The best results are highlighted in \textbf{bold}. The notation (S) indicates supervised methods, (U) denotes unsupervised methods, and (Z) represents Zero-Shot methods, including ours. We compare our method against 6 zero-shot methods, 17 unsupervised methods (12 distinct), and 6 supervised methods. The evaluation of average time (minutes) from three independent runs, memory (Gb), and FLOPs was conducted on the 400 × 600 LOL dataset image using an NVIDIA A10 GPU, under isolated conditions with no concurrent processes. For GDP, the FLOPs measurement exceeded computational limits, indicated as `-'. For RUAS, the NL calculation resulted in `-' due to the presence of few near-white or near-black images. As we leverage decoder component of VAE from QuadPrior~\cite{quadprior} for robust self-reconstruction and a frozen diffusion model, we denote "n/a" for Train Data. Likewise, we apply the same notation for zero-shot diffusion-based methods, including GDP, TAO, and FourierDiff. For notations \hl{(b)}, \hl{(c)}, and \hl{(d)}, please refer to Figure~\ref{fig:taxonomy}.}
\label{tab:quan-metrics}
\end{table*}

\begin{figure*}[ht]
\centering
\includegraphics[width=\textwidth]{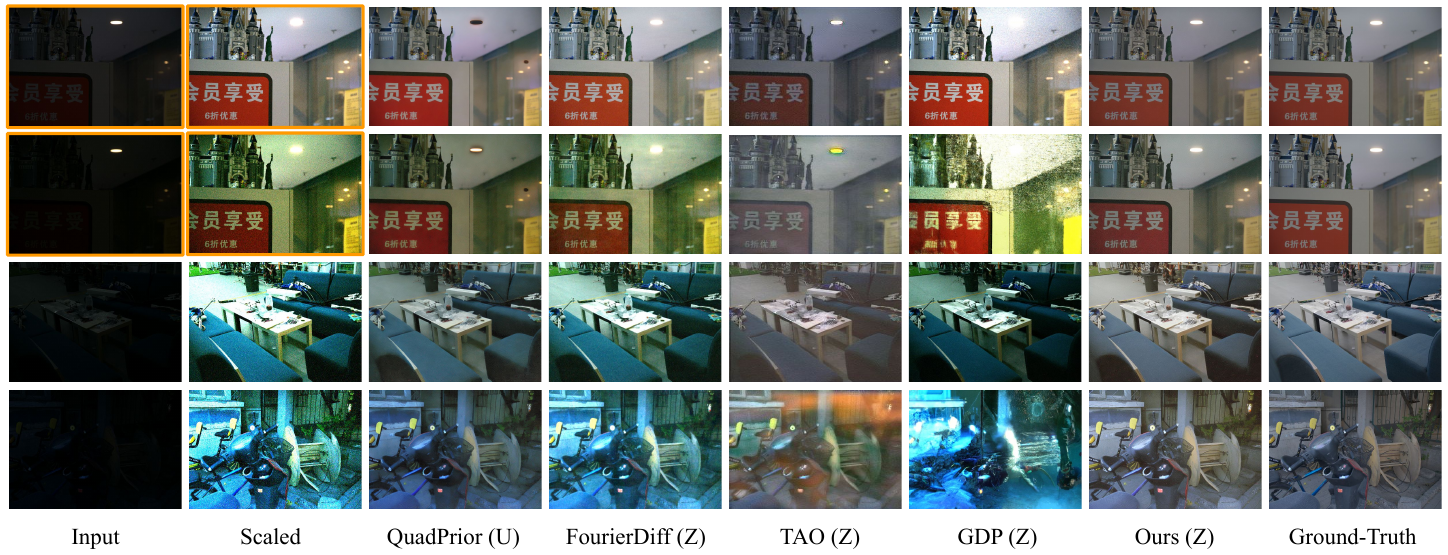} 
\caption{Qualitative evaluation of our method against existing unsupervised and zero-shot approaches on the paired LOL dataset. Please \textbf{zoom in without night-light mode} to accurately compare colors and observe noise reduction in each method. Our method demonstrates consistency with the ground truth as well as across different images of the same scene (see rows 1 and 2), highlighting the reliability and robustness of our approach. Moreover, our method demonstrates reduced susceptibility to incorrect color shifts compared to existing methods and accurately preserves color fidelity.}
\label{fig:qual-paired}
\end{figure*}

\section{Experiments}

\begin{figure*}[ht]
\centering
\includegraphics[width=\textwidth]{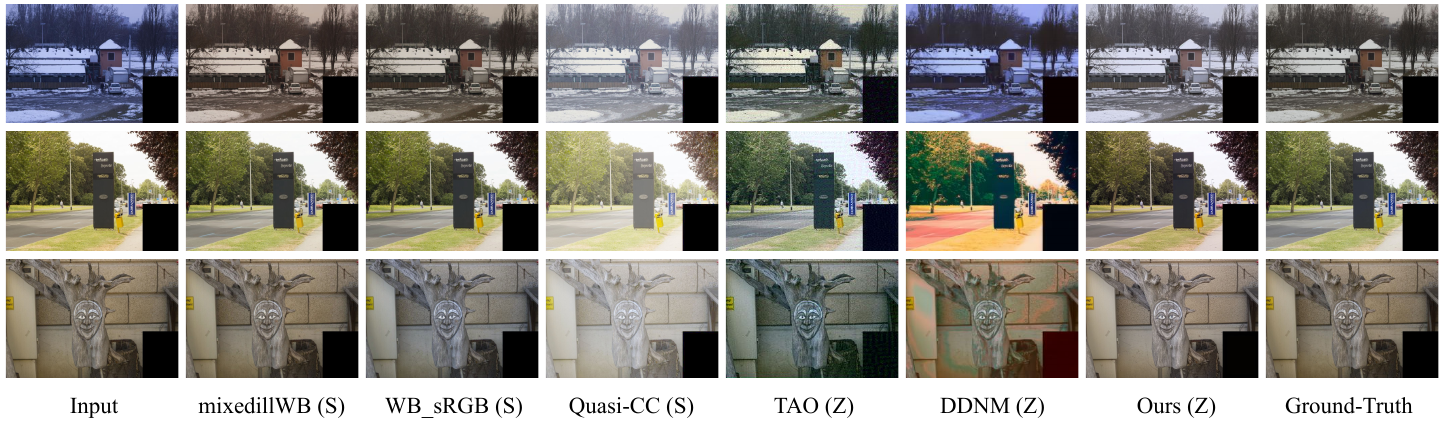} 
\caption{Qualitative evaluation of our method against existing supervised and general image restoration methods on auto white balance task on CUBE+ dataset~\cite{wb_srgb, cube2}. In this dataset, the calibration object is masked out using a black box. Please \textbf{zoom in without night-light mode} to accurately compare colors and observe noise reduction in each method. Without any modifications to the low-light image enhancement framework, our approach achieves competitive performance with supervised AWB methods~\cite{mixedillwb, wb_srgb, quasi-cc} while surpassing state-of-the-art image restoration methods~\cite{ddnm, tao}. The column labeled as Quasi-CC represents the Quasi-CC method trained on the Places365.}
\label{fig:qual-awb}
\end{figure*}

\begin{table}[ht]
\centering
\scriptsize
\setlength{\tabcolsep}{2pt}
\renewcommand{\arraystretch}{1.1}
\begin{tabular}{c|l|c|ccc}
\toprule
& Method & Train Data & $\Delta E \downarrow$ & MAE $\downarrow$ & MSE $\downarrow$ \\
\midrule
 & mixedillWB~\cite{mixedillwb} \textit{WACV'22} & RenderedWB~\cite{wb_srgb} & \textbf{8.03} & \textbf{4.35$^\circ$} & \textbf{118.91} \\
 & WB\_sRGB~\cite{wb_srgb} \textit{CVPR'19} & NUS~\cite{nus}, Gehler~\cite{gehler} & 9.50 & 4.49$^\circ$ & 451.26 \\
S & Quasi-CC~\cite{quasi-cc} \textit{CVPR'19} & Flickr100k~\cite{flickr100k} & 24.44 & 6.84$^\circ$ & 3170.05 \\
 & Quasi-CC~\cite{quasi-cc} \textit{CVPR'19} & ilsvrc12~\cite{ilsvrc12} & 24.64 & 6.85$^\circ$ & 3193.85 \\
 & Quasi-CC~\cite{quasi-cc} \textit{CVPR'19} & Places365~\cite{places365} & 24.15 & 6.58$^\circ$ & 3171.20 \\
\midrule
  & DDNM~\cite{ddnm} \textit{ICLR 2023}   & n/a & 47.76 & 19.97$^\circ$ & 6311.70 \\
Z  & TAO~\cite{tao} \textit{ICLR 2024}   & n/a & 19.06 & 9.81$^\circ$ & 850.05 \\
  & Ours & n/a & \textbf{15.31} & \textbf{6.49$^\circ$} & \textbf{677.89} \\
\bottomrule
\end{tabular}
\caption{Quantitative comparison on the CUBE+ dataset~\cite{wb_srgb, cube2}. We use the exact evaluation code from WB\_sRGB~\cite{wb_srgb}. Metrics include average $\Delta E$ (CIE76), MAE (deg), and MSE. Our method is inherently adaptable to auto white-balance without modifications and outperforms the general image restoration methods TAO~\cite{tao} and DDIM~\cite{ddim} with a degradation function aligning each channel mean. Additionally, our approach achieves competitive results compared to supervised methods trained specifically for this task.}
\label{tab:awb-quan-metrics}
\end{table}

\paragraph{Datasets}  We evaluate the performance of our method on standard benchmarks. For real low-normal paired datasets, we employ the LOL dataset by adopting LOLv1~\cite{lolv1} (15 test images) and LOLv2~\cite{lolv2} (100 test images), along with the LSRW dataset~\cite{lsrw} (50 test images). Additionally, we assess our method on five standard unpaired benchmarks, collectively referred to as Unpaired: DICM~\cite{dicm} (44 low-light images and 20 bright images, totaling 64), LIME~\cite{lime} (total 10 images), NPE~\cite{npe} (total 75 images), MEF~\cite{mef} (17 low-light image sequences with multiple exposure levels, totaling 79), and VV~\cite{vv} (total 24 images). For the Unpaired datasets, we use the exact dataset provided in ~\cite{lolv1}, which is the same source that separately introduces the LOLv1 dataset. Specifically, we report the following image counts: DICM (44 images), LIME (10 images), MEF (79 images), NPE (75 images), and VV (24 images). These numbers are included for precision, as prior studies often omit exact details or show slight discrepancies, particularly for DICM (reported as 44 or 64 images) and NPE (reported as 17, 75, or 84 images in different works). For the paired datasets, we report PSNR, SSIM, LPIPS~\cite{lpips}, and report ILNIQE~\cite{ilniqe}, BRISQUE~\cite{brisque}, and NL~\cite{nl} for the unpaired datasets. For auto white balance (AWB) evaluations, the CUBE+ dataset serves as the basis, where we select the first 200 images out of a total of 10,242. To ensure fairness and reproducibility, the images are ordered by filename, first numerically and then lexicographically, before selection. This method ensures fairness and eliminates selection bias.

\subsection{Experimental Results}
\textbf{Quantitative Results (LLIE).} 
As shown in Table~\ref{tab:quan-metrics}, our approach consistently outperforms existing unsupervised and zero-shot methods across most quantitative metrics. Our method performs on par with supervised methods on the datasets it was trained on while demonstrating superior generalization on datasets that were not seen by supervised methods. On unpaired datasets, our method surpasses all zero-shot, unsupervised, and supervised methods across all metrics, only except for ILNIQE, which demonstrates the highest accuracy among zero-shot methods and the second-highest among all unsupervised methods. On the LOL dataset, our method achieves superior performance in PSNR, SSIM, and LPIPS compared to all zero-shot and unsupervised methods and is on par with supervised methods trained on LOLv1. For the LSRW dataset, our method achieves the best LPIPS score among all zero-shot, unsupervised, and supervised methods, indicating the closest perceptual similarity. Additionally, it achieves the highest PSNR among all zero-shot and unsupervised methods.

\begin{figure*}[t]
  \centering
  \includegraphics[width=\textwidth]{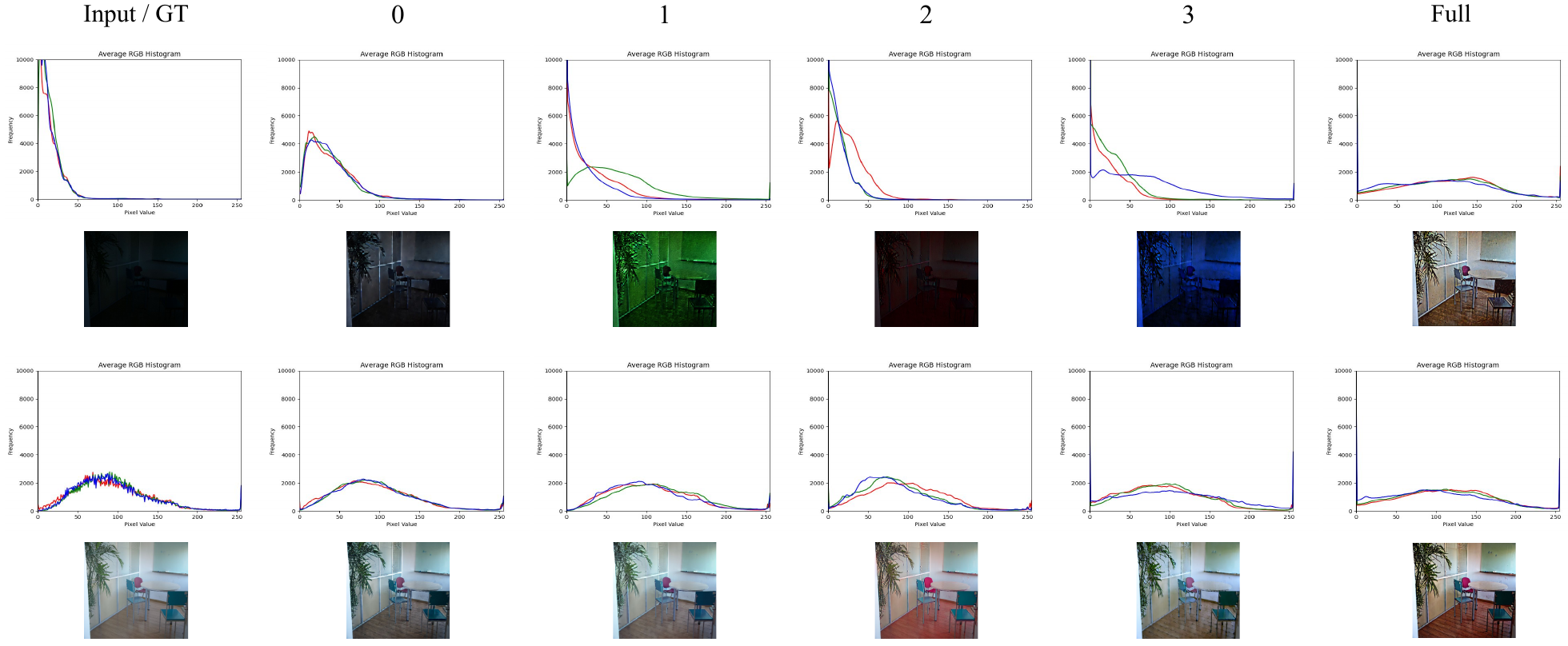}
  \caption{The top row shows the average RGB histograms of 100 randomly selected LOL images for each channel (0, 1, 2, 3) when individually aligned with the Gaussian latent space, as well as when all channels are aligned simultaneously (Full). The image below the histogram provides an example output illustrating this effect. Despite the interdependent nature of the latent space, we hypothesize that each channel exhibits a predominant inclination towards a particular color property. This characteristic, however, is not immediately discernible when analyzing a well-illuminated image, where the color distributions are naturally balanced (row 2). In contrast, when examining darker images, where the red, green, and blue channels are biased towards lower values, these correlations become more conspicuous despite the intertwined nature of the latent space. As a whole, the alignment to the Gaussian latent space subtly shifts and balances colors and luminance, as all channels share a near-zero-centered distribution and symmetry. (row 1, last column). As color degradation is prevalent in both low-light images and color-imbalanced images requiring white balance, this alignment corrects subtle color shifts, a significant benefit in LLIE and AWB tasks.}
  \label{fig:channel-histogram}
\end{figure*}

\begin{figure}[t]
  \centering
  \includegraphics[width=\columnwidth]{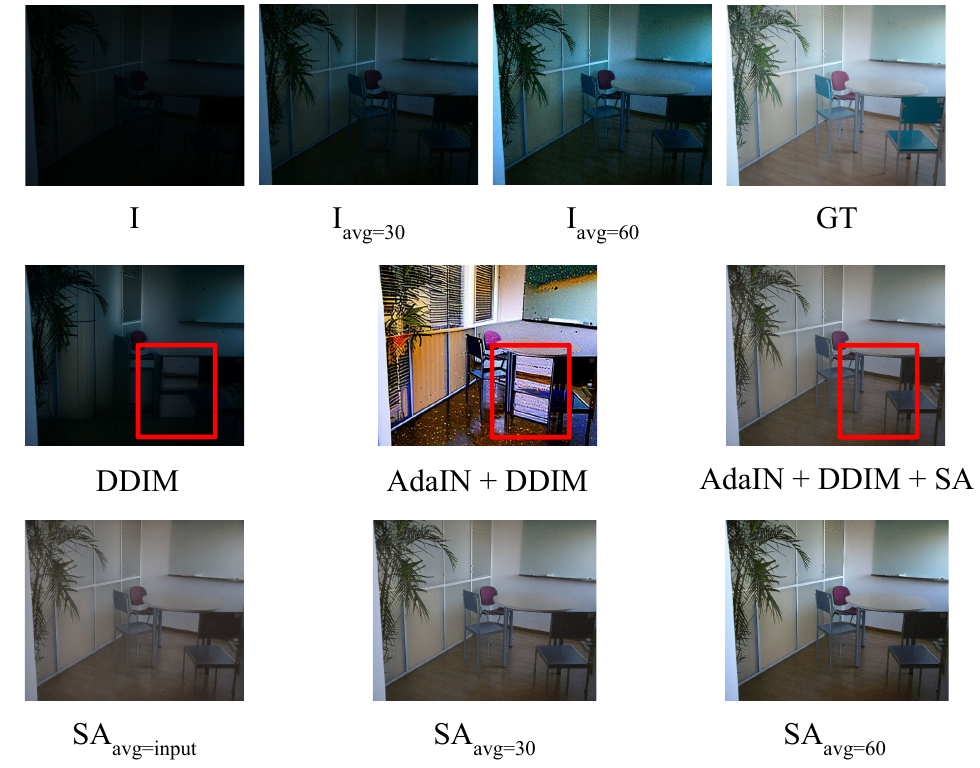}
  \caption{\textbf{Qualitative ablation study.} The images in this figure are sourced from the LOL dataset. Please refer to Section~\ref{sec:ablation_studies} for a detailed discussion.}
  \label{fig:ablation-images}
\end{figure}

\textbf{Quantitative Results (AWB).} 
As illustrated in Table~\ref{tab:awb-quan-metrics}, without any modifications to low-light image enhancement frameworks, our approach achieves competitive performance with supervised AWB methods~\cite{mixedillwb, wb_srgb, quasi-cc} while surpassing state-of-the-art image restoration methods~\cite{ddnm, tao}. 

\textbf{Qualitative Results (LLIE).} 
As illustrated in Figure~\ref{fig:qual-paired}, our method demonstrates consistency with the ground truth as well as across different images of the same scene (see rows 1 and 2), highlighting its reliability and robustness. Moreover, our method exhibits reduced susceptibility to incorrect color shifts and maintains well-balanced brightness while preserving the structural integrity of the images. These qualitative results explain the high-performance metrics achieved by our method. In addition, while TAO~\cite{tao}, a robust image restoration method, exhibits a noticeable color shift, yet this shift does not align with the expected color distribution. This is particularly evident in its LPIPS metrics (where lower values are better) and its metrics in the Unpaired evaluation.

\textbf{Qualitative Results (AWB).}  
As illustrated in Figure~\ref{fig:qual-awb}, although our method is not explicitly trained for the AWB task, its results closely align with the ground truth and is on par with supervised methods~\cite{mixedillwb, wb_srgb, quasi-cc}. In contrast, TAO~\cite{tao} introduces residual noise, DDNM~\cite{ddnm} exhibits color shifts, and Quasi-CC leads to over-exposure.

\section{Ablation Studies}
\label{sec:ablation_studies}
Our framework maintains a simple structure while achieving superior performance over the state-of-the-art in almost every measure. Our analysis suggests three fundamental reasons: \textbf{Inverted layers have meaning:} Despite the inherent interdependence among channels in the latent space, we hypothesize that each channel predominantly aligns with a specific color property as illustrated in Figure~\ref{fig:channel-histogram}. Thus, re-centering an out-of-distribution latent—such as one corresponding to an excessively dark image—toward $\mathcal{N}(0, \mathbf{I})$ through AdaIN~\cite{adain} results in a more balanced reconstruction as shown in Figure~\ref{fig:ablation-images} (row 2, AdaIN + DDIM). As color degradation is prevalent in both low-light image and color-imbalanced images requiring white balance, this alignment correct subtle color shifts. \textbf{Diffusion prior:} Since diffusion models are trained on large-scale datasets of natural images, their reconstruction process is biased towards producing natural images. \textbf{Self-attention features as guidance:} Self-attention features, which are largely invariant to image intensity and white balance, effectively guide the denoising process. The ablation studies presented in Table~\ref{tab:ablation-table} and Figure~\ref{fig:ablation-images} corroborate this interpretation.

\textbf{Ours w/ SA (DDIM Sampling).} Given an input $z_0$, DDIM inversion~\cite{dhariwal2021diffusion, ddim} reverses the DDIM sampling process under the assumption that the underlying ODE can be inverted in the limit of small step sizes:
\begin{equation}
z_{t+1} = \sqrt{\frac{\alpha_{t+1}}{\alpha_t}} z_t + \left(\!\sqrt{\frac{1}{\alpha_{t+1}} - 1} - \sqrt{\frac{1}{\alpha_t} - 1}\!\right) \cdot \epsilon_{\theta} (z_t, t; \varnothing),
\end{equation}%
where \(\alpha_t\) is the noise schedule parameters at diffusion step \(t\), \(z_t\) is the latent state, \(t\) is the timestep, \(\epsilon_{\theta}\) is the diffusion U-Net, and $\varnothing$ as we do not use any text prompt as the conditioning signal.

However, DDIM inversion inherently introduces approximation errors at each time step, leading to failed reconstruction, as the fidelity of the reconstruction is contingent upon the difference between \( z_{t+1} - z_t \). To circumvent this limitation, rather than extracting self-attention features during the sampling phase~\cite{plug_and_play, masactrl, dragdiffusion, pix2pixzero, pix2video}, we instead capture self-attention during the DDIM inversion process as in FateZero~\cite{fatezero}.

\textbf{Ours w/ SD Decoder.} For decoding the final output latent, we use the decoder component of VAE from QuadPrior~\cite{quadprior}, which demonstrates superior self-reconstruction compared to the default Stable Diffusion decoder~\cite{sd1.5}, often prone to distortions from latent compression. 

\textbf{Diffusion internal features.}
Among the internal components of the diffusion model, we leverage self-attention (SA) features, as they demonstrate greater robustness to variations in input compared to residual block features (row 4), as shown in Table~\ref{tab:ablation-table}.

\textbf{Self-Attention.}
In Table~\ref{tab:ablation-table}, Ours w/ SA$_{\text{avg}=input}$ represents deriving self-attention (SA) from an input without preprocessing. Conversely, Ours w/ SA${\text{avg}=60}$, which involves upscaling the input image average intensity to 60 (if it falls below this threshold), not only exhibits more compromised results but also potentially amplifies noise in the image. This suggests that while self-attention remains stable under varying lighting conditions, it is not entirely invariant to illumination changes. 

\begin{table}[ht]
\centering
\resizebox{\columnwidth}{!}{%
\begin{tabular}{|l|c|c|c|}
\hline
\textbf{Model} & \textbf{PSNR $\uparrow$} & \textbf{SSIM $\uparrow$} & \textbf{LPIPS $\downarrow$} \\ \hline
Ours w/ SA (DDIM Sampling) & 20.433 & 0.775 & 0.266 \\ 
\hline
Ours w/ SD Decoder     & 19.927 & 0.600 & 0.248 \\ 
\hline
Ours w/o SA                 & 13.173 & 0.437 & 0.506 \\ 
Ours w/ Res & 19.363 & 0.748 & 0.240 \\ 
\hline
Ours w/ SA$_{\text{avg}=input}$ & 21.238 & 0.822 & 0.187 \\ 
Ours w/ SA$_{\text{avg}=60}$ & 20.798 & 0.785 & 0.194 \\ 
\hline
Ours (final)                 & {\bf 21.739} & {\bf 0.815} & {\bf 0.177} \\ \hline
\end{tabular}%
}
\caption{\textbf{Quantitative ablation study.} The reported metrics are derived from the LOL dataset. Please refer to Section~\ref{sec:ablation_studies} for a detailed discussion.}
\label{tab:ablation-table}
\end{table}

\begin{figure}[t]
  \centering
  \includegraphics[width=\columnwidth]{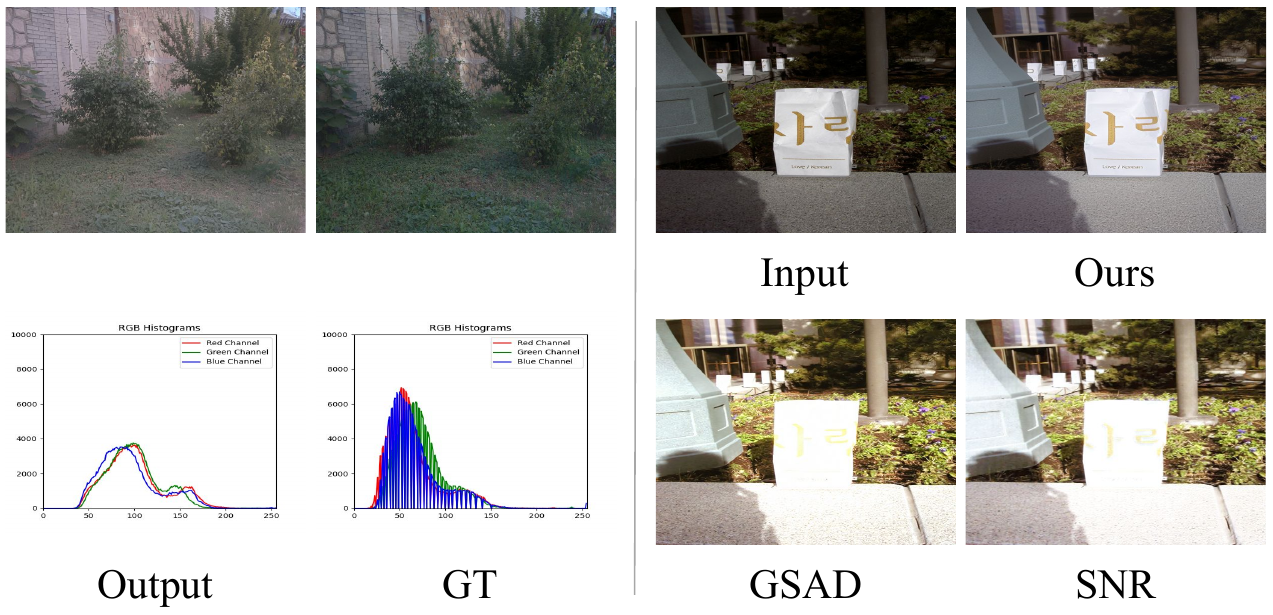}
  \caption{\textbf{Failure cases.} Our method maintains color channel values centered around 100. While this characteristic may occasionally lead to deviations from the ground truth, it proves advantageous in most cases, particularly for low-light images where significant color information is lost. In addition, this property becomes beneficial for inputs with varying brightness, as illustrated on the right. Supervised approaches, GSAD and SNR, trained on LOLv1~\cite{lolv1}, produce overexposed outputs due to their lack of exposure to bright input images. In contrast, our approach demonstrates robustness across varying lighting conditions.}
  \label{fig:failure-cases}
\end{figure}
\section{Discussion and Conclusion}
We introduced a new zero-shot framework for low-light image enhancement, which also serves as the first zero-shot auto white balance method. Our approach requires no training, fine-tuning, or optimization, yet it adheres to the principles of color constancy and achieves superior results than state-of-the-art methods. In contrast to existing zero-shot methods that depend on customized constraints, our method leverages the internal features present in the model itself to guide the inference, and we anticipate that this method will find additional applications in the future.
\clearpage
{
    \small
    \bibliographystyle{ieeenat_fullname}
    \bibliography{main}
}

\end{document}